\documentclass[conference]{IEEEtran}

\usepackage[english]{babel}
\usepackage[utf8x]{inputenc}
\usepackage[T1]{fontenc}
\usepackage[table]{xcolor}
\usepackage{todonotes}
\usepackage{graphicx}
\usepackage{algpseudocode}
\usepackage{algorithm,url}
\usepackage{subcaption}
\usepackage{enumitem,multirow,balance}
\usepackage{amsmath}

\definecolor{Gray}{gray}{0.9}
\definecolor{Gray2}{gray}{0.8}
\definecolor{Gray3}{gray}{0.6}

\newcommand{\bff}[1]{{\textbf{#1}}}

\graphicspath{{./images/}, {./}}

\title{Deep Reinforcement Learning \\for General Video Game AI}
\author{\IEEEauthorblockN{Ruben Rodriguez Torrado*}
\IEEEauthorblockA{New York University\\
New York, NY\\
rrt264@nyu.edu}
\and
\IEEEauthorblockN{
Philip Bontrager*}
\IEEEauthorblockA{New York University\\
New York, NY\\
philipjb@nyu.edu }
\and
\IEEEauthorblockN{Julian Togelius}
\IEEEauthorblockA{New York University\\
New York, NY\\
julian.togelius@nyu.edu }\\
\and
\IEEEauthorblockN{Jialin Liu}
\IEEEauthorblockA{Southern University of Science and Technology\\
Shenzhen, China\\
liujl@sustc.edu.cn}
\and
\IEEEauthorblockN{Diego Perez-Liebana}
\IEEEauthorblockA{Queen Mary University of London\\
London, UK\\
diego.perez@qmul.ac.uk }
}

\begin{document}
\maketitle

\begin{abstract}
The General Video Game AI (GVGAI) competition and its associated software framework provides a way of benchmarking AI algorithms on a large number of games written in a domain-specific description language. While the competition has seen plenty of interest, it has so far focused on online planning, providing a forward model that allows the use of algorithms such as Monte Carlo Tree Search.

In this paper, we describe how we interface GVGAI to the OpenAI Gym environment, a widely used way of connecting agents to reinforcement learning problems. Using this interface, we characterize how widely used implementations of several deep reinforcement learning algorithms fare on a number of GVGAI games. We further analyze the results to provide a first indication of the relative difficulty of these games relative to each other, and relative to those in the Arcade Learning Environment under similar conditions.
\end{abstract}

%\tableofcontents
\section{Introduction}

The realization that video games are perfect testbeds for artificial intelligence methods have in recent years spread to the whole AI community, in particular since Chess and Go have been effectively conquered, and there is an almost daily flurry of new papers applying AI methods to video games. In particular, the Arcade Learning Environment (ALE), which builds on an emulator for the Atari 2600 games console and contains several dozens of games~\cite{bellemare2013arcade}, have been used in numerous published papers since DeepMind's landmark paper showing that Q-learning combined with deep convolutional networks could learn to play many of the ALE games at superhuman level~\cite{mnih2015human}.

As an AI benchmark, ALE is limited in the sense that there is only a finite set of games. This is a limitation it has in common with any framework based on existing published games. However, for being able to test the general video game playing ability of an agent, it is necessary to test on games on which the agent was not optimized. For this, we need to be able to easily create new games, either manually or automatically, and add new games to the framework. Being able to create new games easily also allows the creating of games made to test particular AI capacities.

The General Video Game AI (GVGAI) competitions and framework were created with the express purpose of providing a versatile general AI benchmark~\cite{ebner2013towards,perez20162014,perez2016general,perez2018general}. The \emph{planning tracks} of the competition, where agents are given a forward model allowing them to plan but no training time between games, have been very popular and seen a number of strong agents based on tree search or evolutionary planning submitted. A \emph{learning track} of the competition has run once, but not seen many strong agents, possibly because of infrastructure issues. For the purposes of testing machine learning agents (as opposed to planning agents), GVGAI has therefore been inferior to ALE and similar frameworks.

In this paper, we attempt to rectify this by presenting a new infrastructure for connecting GVGAI to machine learning agents. We connect the framework via the OpenAI Gym interface, which allows the interfacing of a large number of existing reinforcement learning algorithm implementations. We plan to use this structure for the learning track of the GVGAI competition in the future. In order to facilitate the development and testing of new algorithms, we also provide benchmark results of three important deep reinforcement learning algorithms over eight dissimilar GVGAI games.

\section{Background}

\subsection{General Video Game AI}

The General Video Game AI (GVGAI) framework is a Java-based benchmark for General Video Game Playing (GVGP) in 2-dimensional arcade-like games~\cite{perez2016general}. This framework offers a common interface for bots (or agents, or controllers) and humans to play any of the more than $160$ single- and two-player games from the benchmark. These games are defined in the Video Game Description Language (VGDL), which was initially proposed by Ebner \textit{et al.}~\cite{ebner2013towards} at the Dagstuhl Seminar on Artificial and Computational Intelligence in Games. 

VGDL~\cite{schaul2013video} is a game description language that defines 2-dimensional games by means of two files, which describe the game and the level respectively. The former is structured in four different sections, detailing game sprites present in the game (and their behaviors and parameters), the interactions between them, the termination conditions of the game and the mapping from sprites to characters used in the level description file. The latter describes a grid and the sprite locations at the beginning of the game. These files are typically not provided to the AI agents, who must learn to play the game via simulations or repetitions. More about VGDL and sample files can be found on the GVGAI GitHub project\footnote{\url{https://github.com/EssexUniversityMCTS/gvgai/wiki/VGDL-Language}}.

The agents implement two methods to interact with the game: a constructor where the controller may initialize any structures needed to play, and an \texttt{act} method, which is called every game frame and must return an action to execute at that game cycle. As games are played in real-time, the agents must reply within a time budget (in the competition settings, $1$ second for the constructor and $40$ms in the \texttt{act} method) not to suffer any penalty. Both methods provide the agent with some information about the current state of the game, such as its status (if it is finished or still running), the player state (health points, position, orientation, resources collected) and anonymized information about other sprites in the game (so their types and behaviours are not disclosed). Additionally, controllers also receive a forward model (in the \textit{planning} setting) and a screen-shot of the current game state (in the \textit{learning} setting).

The GVGAI framework has been used in a yearly competition, started in 2014, and organized around several tracks. Between the single-~\cite{perez20162014} and the two-player~\cite{gaina2017gvgai} GVGAI planning competitions, more than $200$ controllers have been submitted by different participants, in which agents have to play in sets of $10$ unknown games to decide a winner. These tracks are complemented with newer ones for single-player agent learning~\cite{liu2017learningmanual,perez2018general}, level~\cite{khalifa2016general} and rule generation~\cite{khalifa2017rulegen}. Beyond the competitions, many researchers have used this framework for different types of work on agent AI, procedural content generation, automatic game design and deep reinforcement learning, among others~\cite{perez2018general}.

In terms of learning, several approaches have been made before the single-player learning track of the GVGAI competition was launched. The first approach was proposed by Samothrakis et al.~\cite{samothrakis2015neuroevolution}, who implemented Separable Natural Evolution Strategies (S-NES) to evolve a state value function in order to learn how to maximize victory rate and score in $10$ games of the framework. Samothrakis et al.~\cite{samothrakis2015neuroevolution} compared a linear function approximator and a neural network, and two different policies, using features from the game state.

Later, Braylan and Miikkulainen~\cite{braylan2016object} used logistic regression to learn a forward model on $30$ games of the framework. The objective was to learn the state (or, rather, a simplification consistent of the most relevant features of the full game state) that would follow a previous one when an action was supplied, and then apply this model in different games, assuming that some core mechanics would be shared among the different games of the benchmark. Their results showed that these learned object models improved exploration and performance in other games.

More recently, Kunanusont et al.~\cite{kunanusont2017general} interfaced the GVGAI framework with DL4J\footnote{Deep Learning for Java: https://deeplearning4j.org/} in order to develop agents that would learn how to play several games via screen capture. $7$ games were employed in this study, of increasing complexity and screen size and also including both deterministic and stochastic games. Kunanusont et al.~\cite{kunanusont2017general} implemented a Deep Q-Network for an agent that was able to increase winning rate and score in several consecutive episodes. 

The first (and to date, only) edition of the single-player learning competition, held in the IEEE’s 2017 Conference on Computational Intelligence in Games (CIG2017), received few and simple agents. Most of them are greedy methods or based on Q-Learning and State-Action-Reward-State-Action (SARSA), using features extracted from the game state. For more information about these, including the final results of the competition, the reader is referred to~\cite{perez2018general}.

\subsection{Deep Reinforcement Learning}
A Reinforcement Learning (RL) agent learns through trial-and-error interactions with a dynamic environment~\cite{sutton1998reinforcement} and balance the reward trade-off between long-term and short-term planning. RL methods have been widely studied in many disciplines, such as operational research, simulation-based optimization, evolutionary computation and multi-agent system, including games. The cooperation between the RL methods and Deep Learning (DL) has led to successful applications in games. More about the work on Deep Reinforcement Learning till 2015 can be found in the review by J. Schmidhuber~\cite{schmidhuber2015deep}. For instance, Deep Q-Networks has been combined with RL to play several Atari 2600 games with video as input~\cite{mnih2013playing,mnih2015human}. Vezhnevets~et al.\cite{vezhnevets2016strategic} proposed STRategic Attentive Writer-exploiter(STRAWe) for learning macro-actions and achieved significant improvements on some Atari 2600 games. 
\emph{AlphaGo}, combined tree search with deep neural networks to play the game of Go and self-enhanced by self-playing, is ranked as 9 dan professional~\cite{silver2016mastering} and is the first to beat human world champion of Go.
Its advanced version, \emph{AlphaGo Zero}~\cite{silver2017mastering} is able to learn only by self-playing (without the data of matches played by human players) and outperforms \emph{AlphaGo}.

During the last few years, several authors have improved the results and stability obtained with the original Deep Q-Networks paper. Wang et. al. \cite{wang2016dueling} introduces a new architecture for the networks know as dueling network, this new architecture uses two separate estimators: one for the state value function and one for the state-dependent action advantage function. The main benefit of this factoring is to generalize learning across actions without imposing any change to the underlying reinforcement learning algorithm. 

Mnih et. al., in 2016, successfully applied neural networks to actor-critic RL \cite{mnih2016asynchronous}. The network is trained to predict both a policy function and a value function for a state, the actor and the critic. Asynchronous Advantage Actor-Critic, A3C, is inherently parallelizable and allows for a big speedup in computation time. The interaction between the policy output and the value estimates has been shown to be relatively stable and accurate for neural networks. This new approach increases the score obtained from the original DQN paper, reducing the computational time by half even without using CPU.

%Focus on DQN %\cite{mnih2013playing}, %\cite{mnih2015human}, additions %\cite{schaul2015prioritized} , and %Actor Critic %\cite{mnih2016asynchronous}

\subsection{OpenAI Gym}

RL is a hot topic for the research community of artificial intelligence. Recent advances that combine DL with RL (Deep Reinforcement Learning) have shown that model-free optimization, or policy gradients, can be used for complex environments. However, in order to continue testing new ideas and increasing the quality of results, the research community needs good benchmark platforms to compare results. This is the main goal of OpenAI GYM platform \cite{brockman2016openai}.

The OpenAI GYM platform provides a high variety of benchmark, such as Arcade Learning Environment (ALE) \cite{arcade2013}, which is a collection of Atari 2600 video games. OpenAI Gym has more environments for testing RL in different types of environments. For example, MuJoCo is used to test humanoid like movement in 2D and 3D. 

\section{Methods}

While one of the main benefits for GVGAI is the ease to which new games can be created for a specific problem, we also feel it is necessary to place the current GVGAI games in the context of other existing environments. This serves two purposes: it further demonstrates the strengths and weaknesses of the current generation of reinforcement learning algorithms, and it allows results achieved on GVGAI to be compared to other existing environments.

\subsection{GVGAI-OpenAI embedding}
The learning competition is based on the GVGAI framework, but no forward model is provided to the agents, thus no simulations of a game are accessible.
However, an agent still has access to the observation of current game state, a \emph{StateObservation} object, provided as a Json object in \emph{String} or as a screen-shot of the current game screen (without the screen border) in \emph{png} format. 
At every game tick, the server sends a new game state observation to the agent, the agent returns either an action to play in $40$ms or requests to abort the current game. When a game is finished or aborted, the agent can select the next level to play, among the existing levels (usually 5 levels). 
This setting makes it possible to embed the GVGAI framework as an OpenAI Gym so that the reinforcement learning algorithms can be applied to learn to play the GVGAI games. Thanks to VGDL, it is easy to design and add new games and levels to the GVGAI framework.

The main framework is described in the manual by Liu et al.~\cite{liu2017learningmanual}, as well as the default rules in the framework. Only 5 minutes is allowed to each of the agents for learning. It is notable that only the decision time (no more than $40ms$ per game tick) used by the agent is included, while the game advancing time, game state serialization time and communication time between the client and agent are not included. The real execution of the learning phase can last several hours.

\subsection{GVGAI Games}

%-------------------------------------------------------------
\begin{figure}[h]
\centering
\includegraphics[width=.8\linewidth]{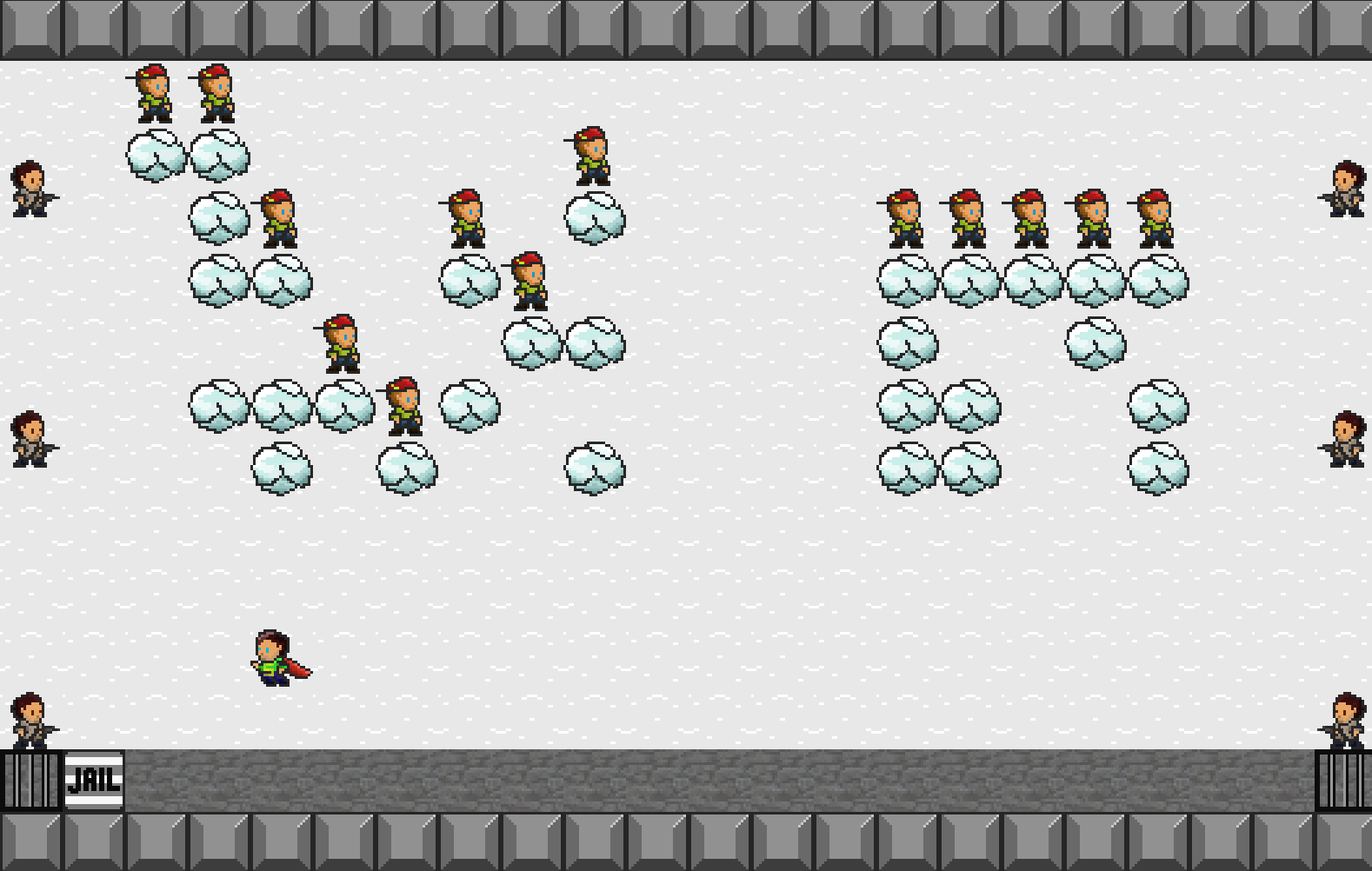}
\caption{\label{fig:gamesuperman}Screenshot of game \emph{Superman}. In this game, innocent civilians are standing on clouds while malicious actors spawn around the edge of the screen and attempt to shoot the clouds out from underneath them. If all the clouds are gone the civilian will fall and only Superman can save them by catching them for 1 point. Superman can also jail the villains for 1 point. If Superman catches all the villains, the player wins and earns an additional 1000 points.} 
\end{figure}
%-------------------------------------------------------------

The GVGAI environment currently has over 160 games and counting. To showcase the environment and the challenges that already exist we sample a number of games to benchmark against popular reinforcement learning algorithms.

Our criteria for sampling games was informal but based on several considerations. Since many of the games in the GVGAI framework have been benchmarked with planning agents, we can roughly rank the games based on how difficult these games are for planning. We tried to get an even distribution across the range going from games that are easy for planning agents, like Aliens, to very difficult, like Superman. The game difficulties are based on the analysis by Bontrager et al. \cite{bontrager2016matching}. Other things we considered were having a few games that also exist in Atari for some comparison and including games that we believed would provide interesting challenges to reinforcement learning agents. Some games in VGDL contain stochastic components as well, mostly in the form of NPC movement. GVGAI has five levels for each game, we used the first level for each game for all the training.

\begin{figure}[h]
\centering
\includegraphics[width=.6\linewidth]{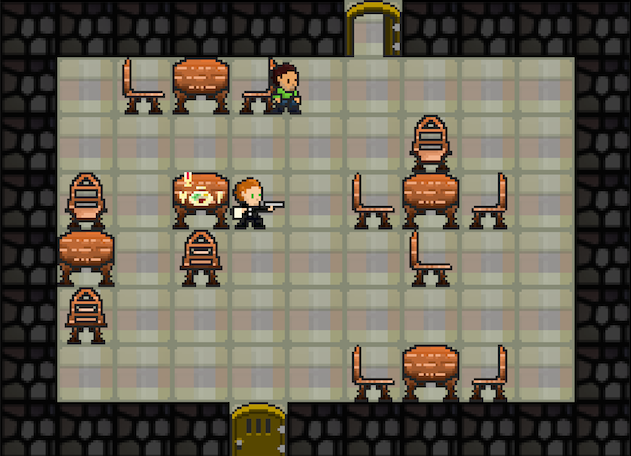}
\caption{\label{fig:gamewaitforbreakfast}Screenshot of game \emph{Wait For Breakfast}. In this game, all tables are empty when a game starts. At a randomly selected game tick, a waiter (in black) serves a breakfast to the table with only one chair. The player (in green) wins the game only if it sits on the chair on the table after the breakfast is served and eats it. The player loses the game if it leaves the chair once breakfast has been served without eating it.} 
\end{figure}

We settled on Aliens, Seaquest, Missile Command, Boulder Dash, Frogs, Zelda, Wait For Breakfast, and Superman. The first five mentioned are modeled after their similarly named Atari counterpart. Zelda consists of finding a target while killing or avoiding enemies. Frogs is modeled after Frogger which is also similar to the Atari Freeway game. Wait For Breakfast (Figure \ref{fig:gamewaitforbreakfast}) is a strange game where the player must go to a breakfast table where food is being served a sit there for a short amount of time. This is not usually what people think of as a game but provides an interesting challenge for bots. Finally, Superman (Figure \ref{fig:gamesuperman}) is a complicated game that involves saving people in a dangerous environment with no reward until the person is safe. A full version of our implementation can be found on GVGAI GYM repository \footnote{\url{https://github.com/rubenrtorrado/GVGAI_GYM}}.

%-------------------------------------------------------------
%-------------------------------------------------------------

\subsection{Benchmarks}

To have standardized results we decided to choose a few popular reinforcement learning algorithms that are provided by the OpenAI Gym baselines library. The baselines are open implementations of these algorithms and are closely based on the original papers \cite{baselines}. The hope is that by using publicly vetted and accessible code that our results will be comparable to other work and reproducible.

From OpenAI’s baseline library we selected three algorithms: Deep Q-Networks (DQN), Prioritized Dueling DQNs, and Advantage Actor-Critic (A2C). These were chosen in part because they have been well documented in similar environments such as ALE. DQN and A3C, which A2C is based on, are the baseline for which many new RL developments are scored against. For this reason, we felt it made sense to use these to benchmark the GVGAI games.

For all three baselines, we used the same network first described in Mnih et al. for playing Atari \cite{mnih2013playing}. This consists of 3 convolutional layers and two fully connected layers as seen in Table \ref{tab:network}. GVGAI is providing screen-shots for each game state that the convolutional network learns to interpret. Each algorithm is trained on one million frames of a particular game. From initial testing, it appeared that one million calls were enough to give an indication of the difficulty of a game for our agents while also being realistic in terms of computational resources. It is also a step in the right direction for the learning track of GVGAI where there are very tight time constraints. To accommodate the smaller number of training iterations, we changed a few training parameters. Buffer size, the size of replay memory, was set to 50,000, the network starts learning after only 1000 initial decisions, and the target Q-network gets updated every 500 steps.

We test both the original DQN and a modified DQN. OpenAI Baselines has a DQN implementation that is based on the original DQN but it also offers prioritized experience replay and dueling networks as options that can be turned on since they work together with the original implementation \cite{baselines}. We tested the original for comparisons and also ran DQN with the two additional modifications to get results from a more state of the art DQN. We used the baseline defaults for the network with a couple of exceptions pertaining to training time. The defaults have been tuned for ALE and should carry over.

To test A3C, OpenAI provides A2C. This is a synchronous version that they found to be more efficient and perform just as well on Atari \cite{baselines}. This was also tested with the baseline defaults with the same changes made for DQN. Each baseline was tested on every game for one million calls, resulting in a total of 24 million calls.

%-------------------------------------------------------------
 \begin{table}[!t]
 \centering
 \setlength\tabcolsep{1.5pt}
 \begin{tabular}{|c|ccc|}
 \hline
 \multirow{2}{*}{\bff{Layer Type}} &\multicolumn{3}{c|}{\bff{Layer Parameters}}\\
  & \bff{Depth}& \bff{Kernel} & \bff{Stride}\\
 \hline
 Convolution 1 & 32 & 8 & 4\\ 
 Convolution 2 & 64 & 4 & 2\\  
 Convolution 3 & 64 & 3 & 1\\  
 Fully Connected & 256 & &\\   
 Fully Connected & Action Space & & \\   
\hline
\end{tabular}
\caption{This table represents the architecture of the network used to play each game. For convolutional layers, depth refers to the convolutional filters and for the fully connected layers it refers to the output size.}
\label{tab:network}
\end{table}
%-------------------------------------------------------------

\section{Results and Discussion}

Here we present the results of training the baselines on each game. The results show the performance of the provided baselines for a sample of the games in the GVGAI framework. This provides insight into how the baselines compare to other AI techniques and to how the GVGAI environment compares to other environments. 

Finally, this section is structured in three parts. First, the results of training the learning algorithms on the games are provided with some additional qualitative remarks. Second, the GVGAI environment is compared to the Atari environment. Third, the reinforcement agents are compared to planning agents that have been used within the framework. 

\subsection{Results of learning algorithms}

\begin{figure*}
  \begin{subfigure}{\linewidth}
	\includegraphics[width=.25\linewidth]{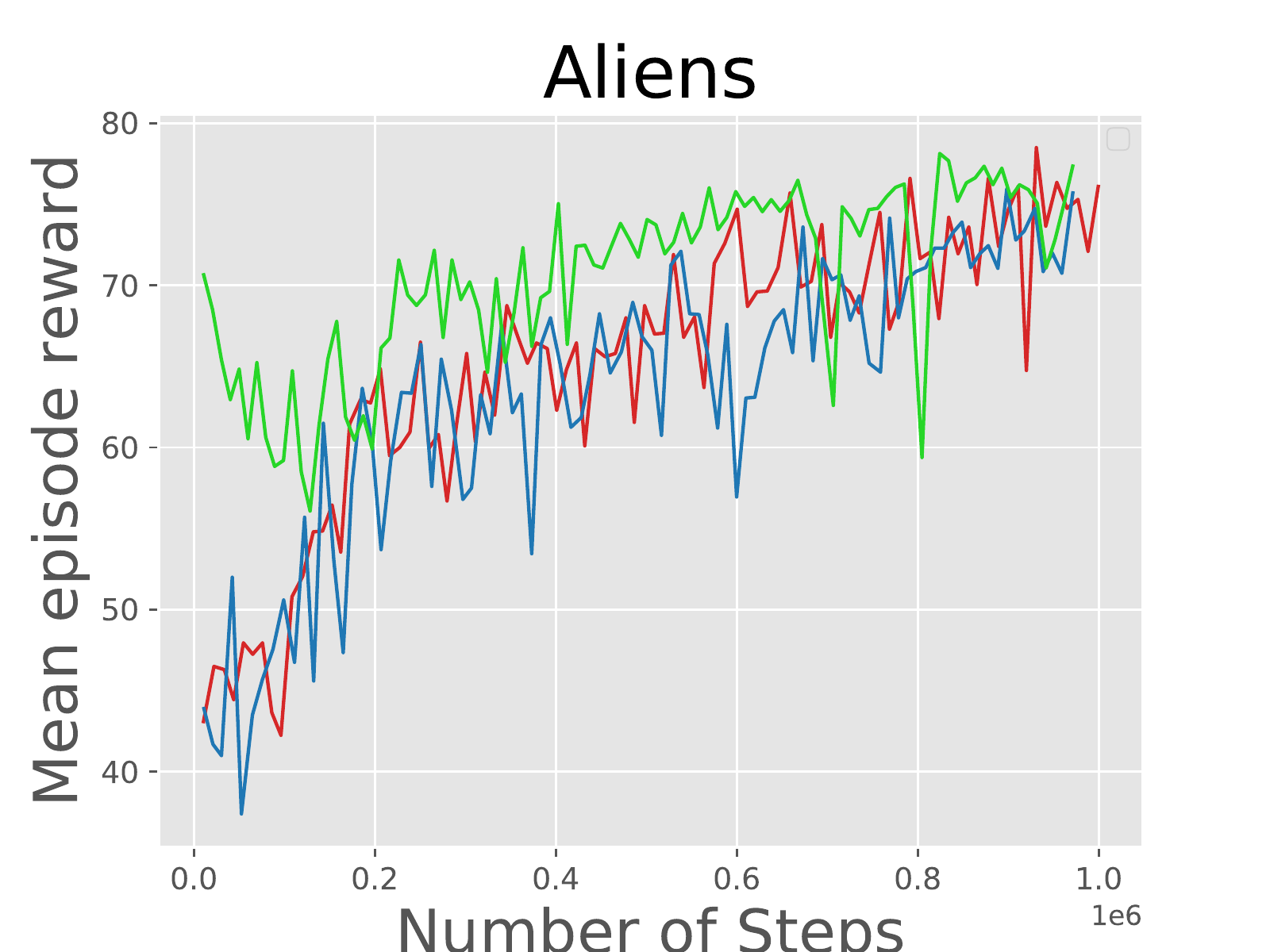}\hfill
 	\includegraphics[width=.25\linewidth]{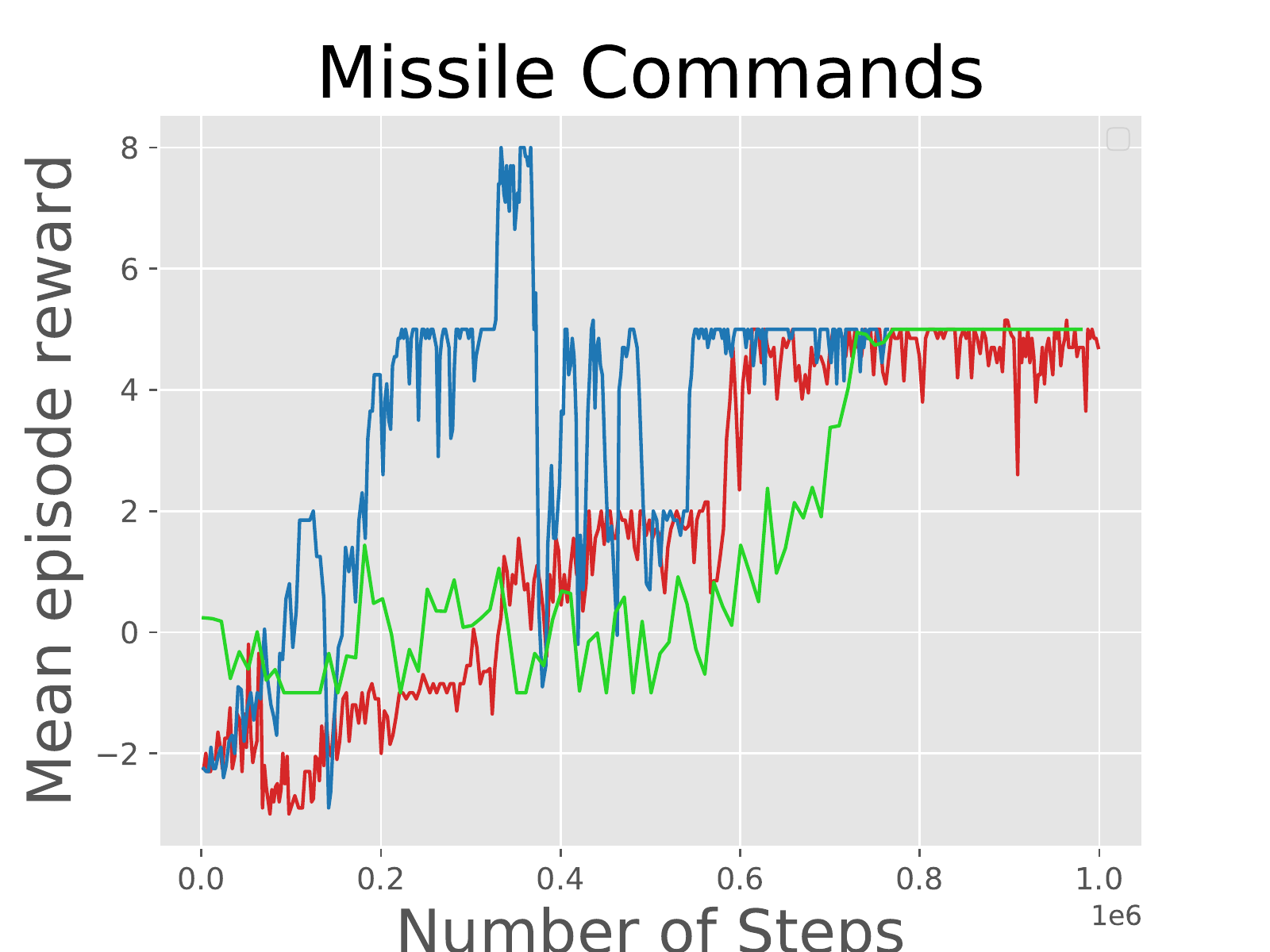}\hfill
 	\includegraphics[width=.25\linewidth]{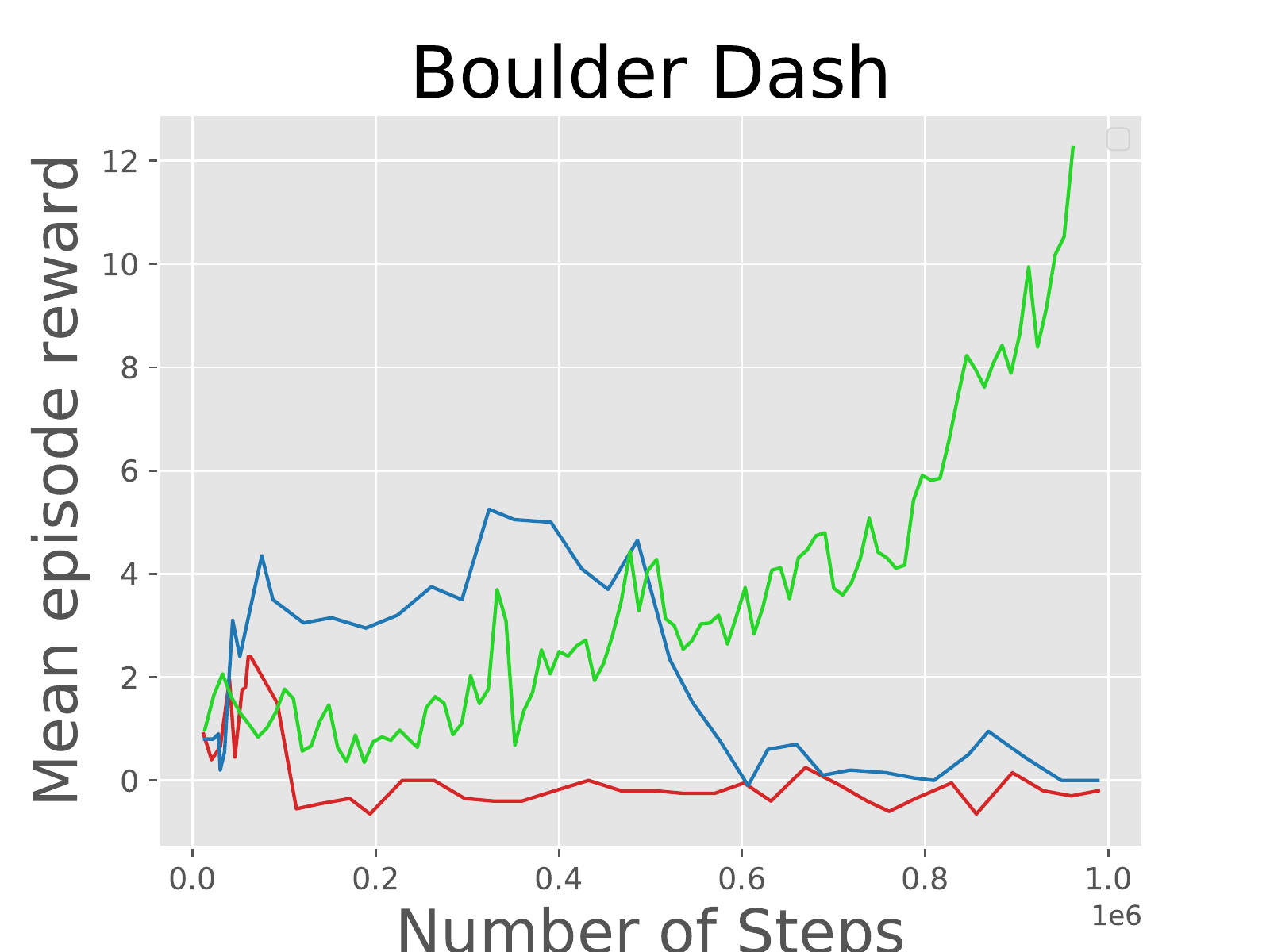}\hfill
    \includegraphics[width=.25\linewidth]{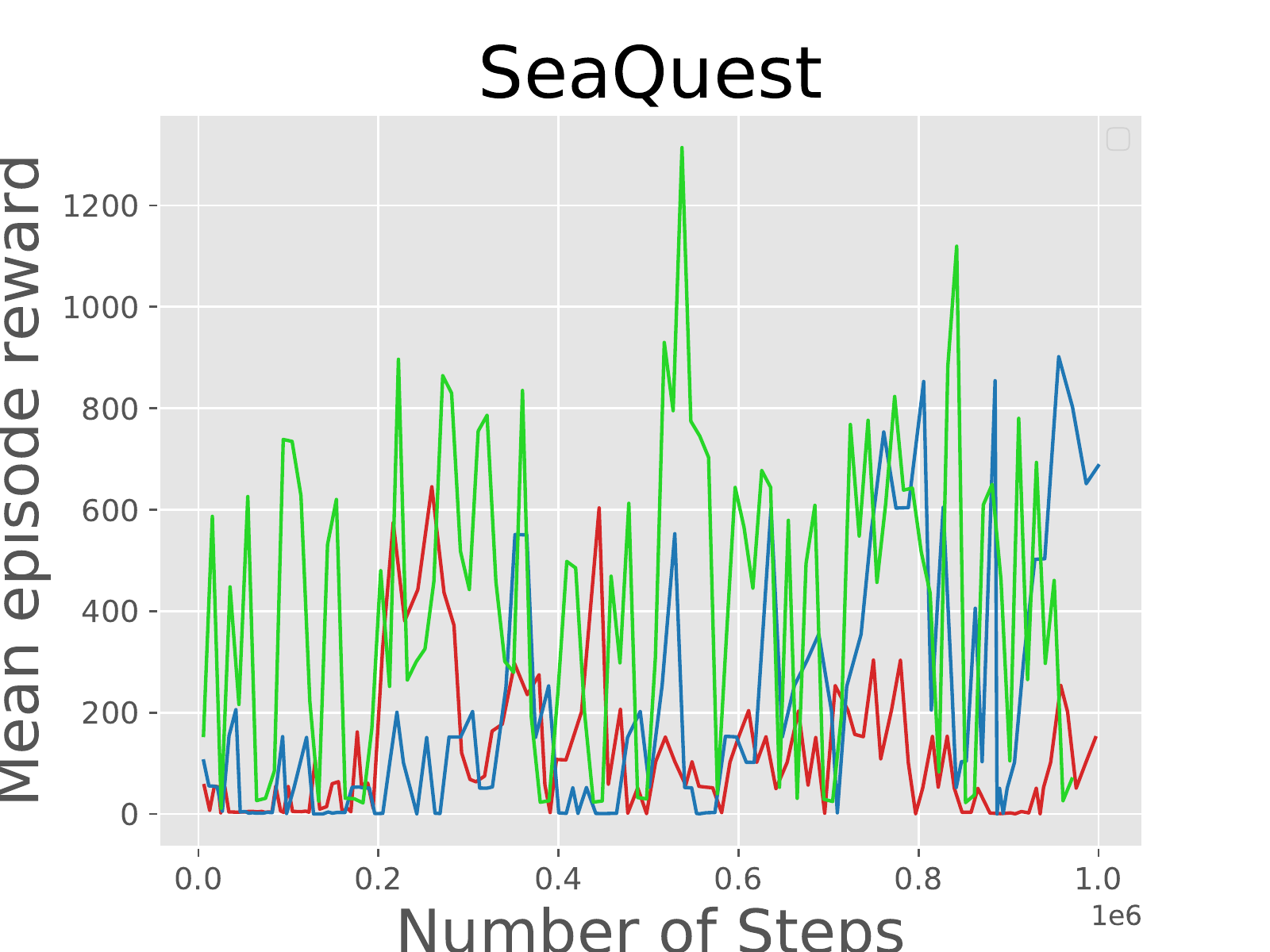}\hfill
  \end{subfigure}\par\medskip
  \begin{subfigure}{\linewidth}
 	\includegraphics[width=.25\linewidth]{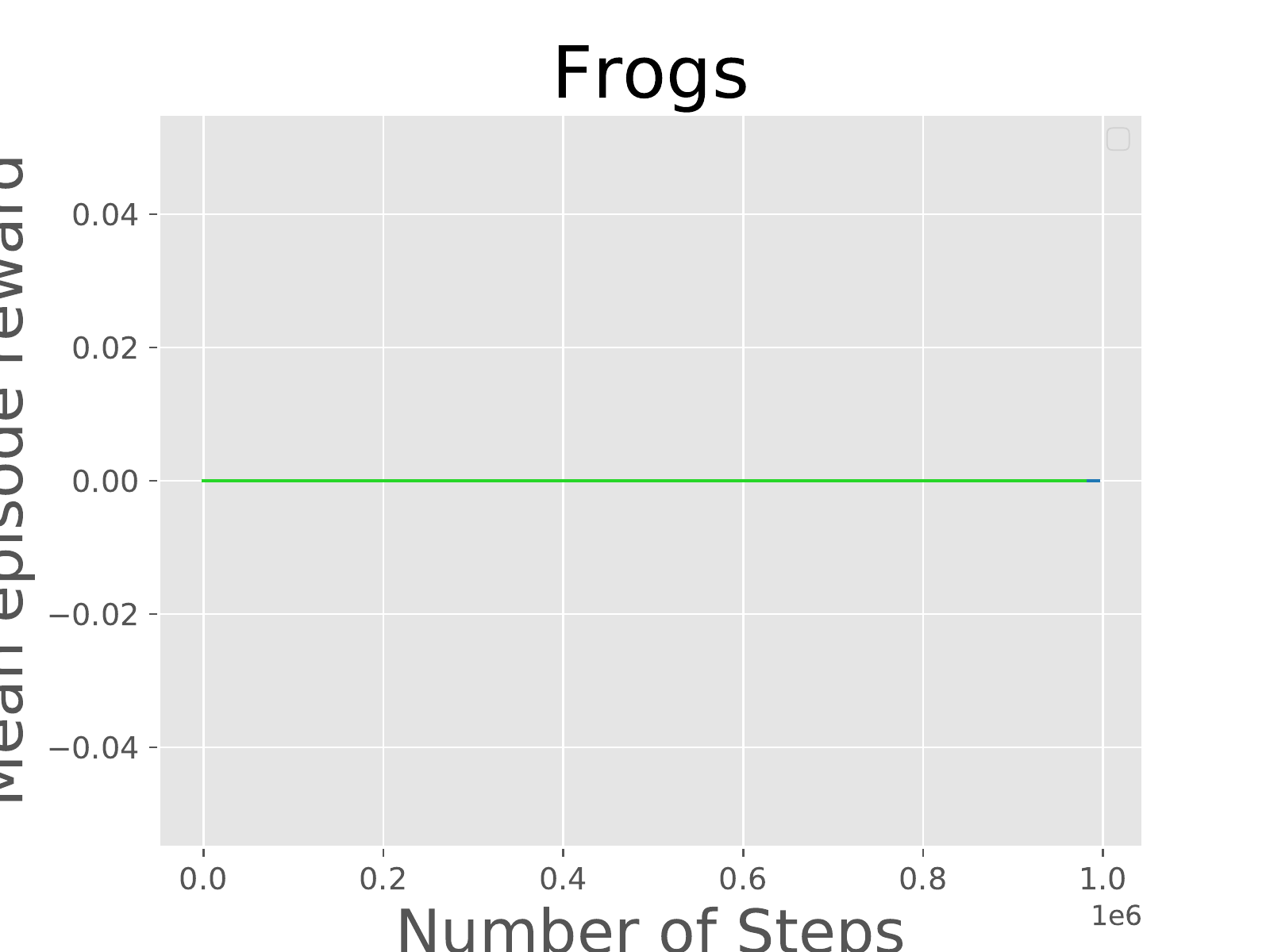}\hfill
 	\includegraphics[width=.25\linewidth]{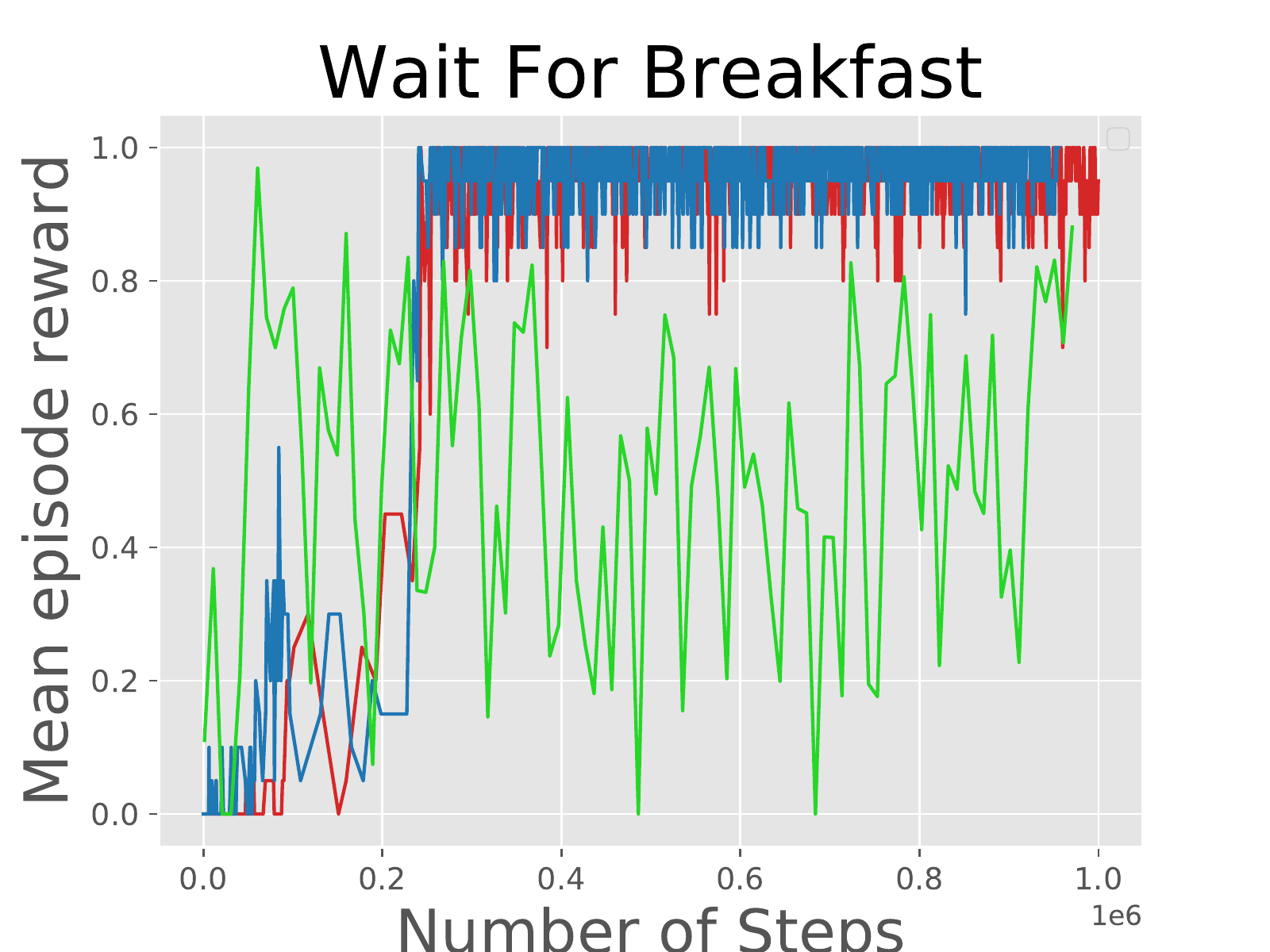}\hfill
    \includegraphics[width=.25\linewidth]{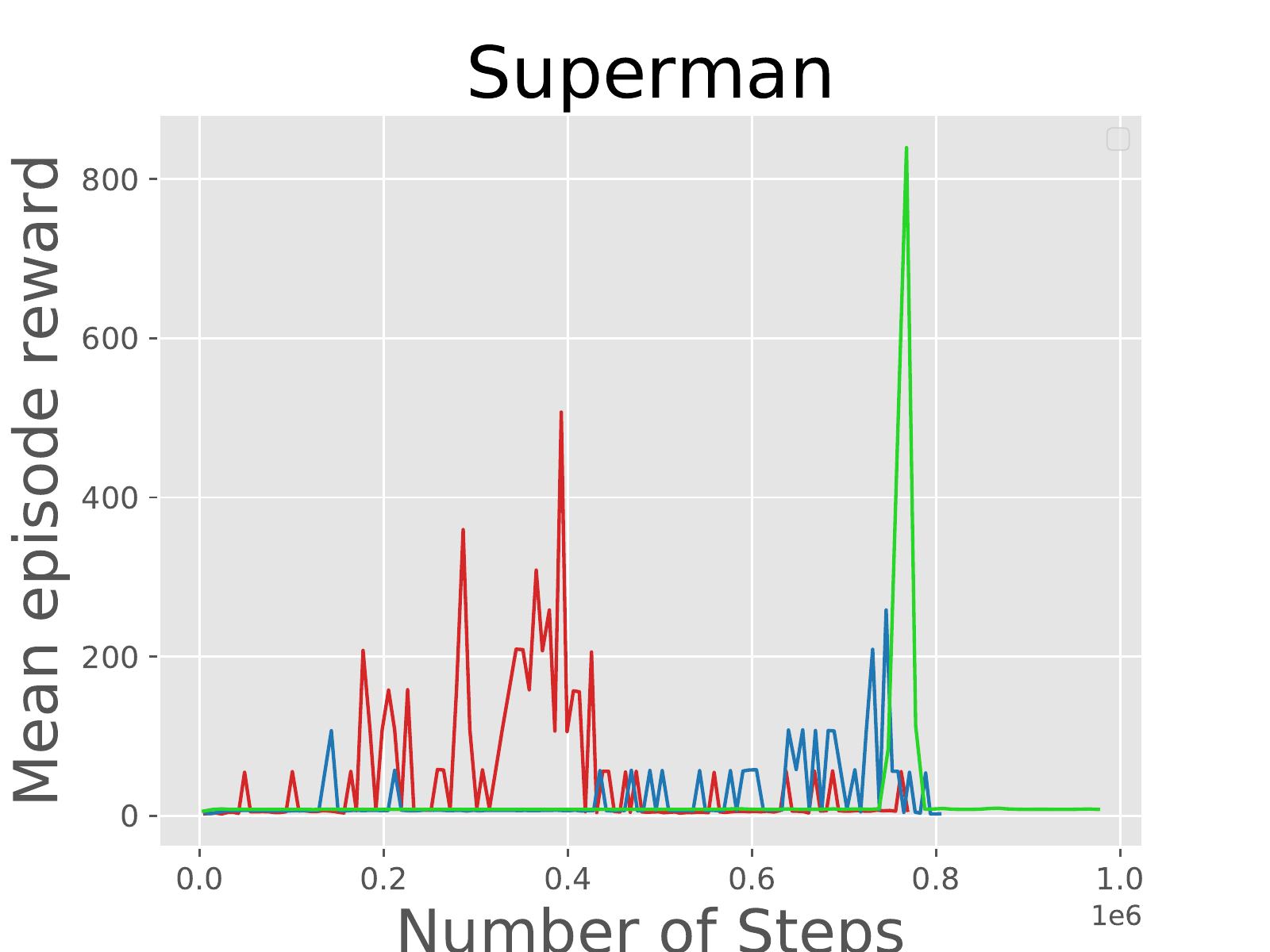}\hfill
 	\includegraphics[width=.25\linewidth]{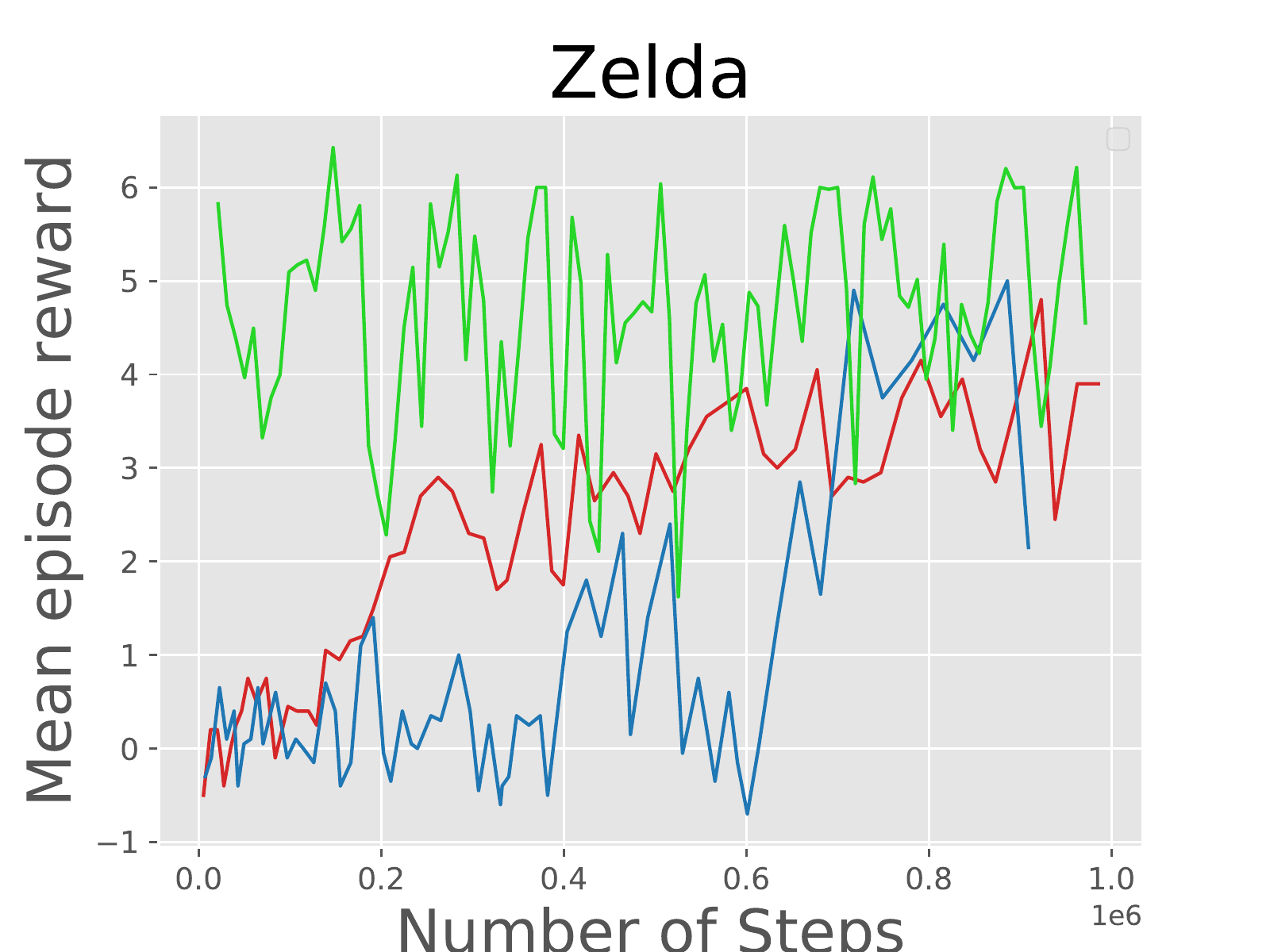}\hfill
  \end{subfigure}\par\medskip
  \caption{Training reward for DQN (red), Prioritized Dueling DQN (blue), and A2C (green). The reward is reported on the y-axis and is different for each game. As an example, Frogs only returns a score of 1 for winning and 0 otherwise. Each algorithm is trained on one million game frames.}
  \label{fig:results}
\end{figure*}

\begin{figure}[h]
\centering
\includegraphics[width=.9\linewidth]{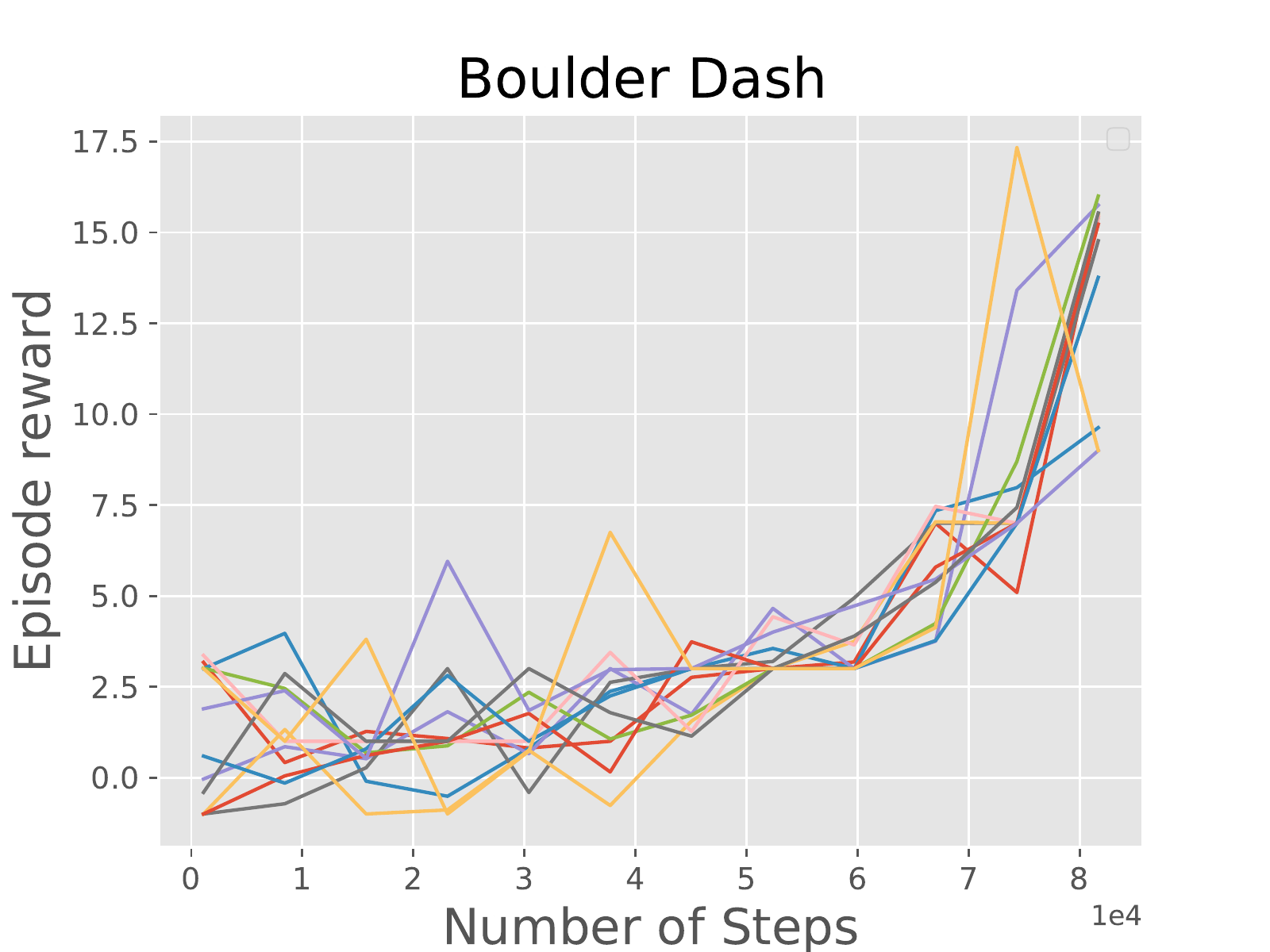}
\caption{\label{fig:a2c} Training reward for all 12 workers of A2C learning on Boulder Dash} 
\end{figure}

Figure \ref{fig:results} shows the training curves for DQN (red), Dueling Prioritized DQN (blue) and A2C (green). The graphs show the total rewards for playing up to that point in time. Rewards are completely defined by the game description so they can't be compared between different games. This is done by reporting the sum of the incremental rewards for the episode at a given time step. Since this data is noisy due to episode restarts, the 20 results are averaged to smooth the graph and better show a trend. A2C allows running in parallel, we were able to run 12 networks in parallel at once. To keep the comparisons fair, A2C is still only allowed one million GVGAI calls and therefore each of the 12 networks is given one-twelfth of a million calls each. This results in the training graph seen in Figure \ref{fig:a2c}. To compare this with the linear algorithms, each time step of A2C is associated with 12 time-steps of the DQN algorithms in Figure \ref{fig:results}. The value for each time step of A2C is the average of all 12 rewards.

Due to the fact that we are running experiments on different machines with different GPU and CPU configurations, we align the results on iterations instead of time. It is important to note that since A2C runs its fixed number of GVGAI calls in parallel, it runs at about 5x the speed of DQN on a machine with two NVIDIA Tesla k80 GPUs. 

Figure \ref{fig:a2c} shows the training curve in parallel for A2C on Boulder Dash. The individual agents are chaotic which helps A2C  break out of local minima. This also points to the importance of the exploration algorithm in learning to play games. In Boulder Dash, as long as one of the 12 workers found an improvement they would all gain. 

The agents were able to learn on most of the games that were sampled. A2C performed the best for most of the games tested. Though it's important to remember a relatively small computational budget was allowed for these algorithms and the others might eventually catch up. 8 games is also a small sample for comparing which algorithm is the best. A2C seems to benefit from sampling more initial conditions and starts with a higher score. 

DQN and Prioritized Dueling DQN were both given the same initial seed so they had the same initial exploration pattern. For this reason, both algorithms tended to start out with similar performance and then diverge as time goes on. Prioritized Dueling DQN seems to slightly outperform vanilla DQN, but on overall they are very similar. A2C could not be compared in this way as it intentionally is running different explorations in parallel and then learn from all of them at the same time. This can explain why A2C tends to start out better right from the beginning, especially in Aliens. It is benefiting from 12 different initial conditions in this case.

Available rewards have a big impact on the success of RL and that is not different in the GVGAI environment. The games where the agents performed worst were the games that had the least feedback. For this work, we left the games in their current form, but it is very easy for researchers to edit the VGDL file and modify the reward structure to create various experiments.

The games sampled here vary a lot in terms of the rewards they offer. Frogs and Wait For Breakfast only provide a single point for winning. This is evident in their training graphs. For Frogs, none of the agents appear to have found a winning solution in the calls allotted. This resulted in a situation where RL could not play the game. Wait For Breakfast has a simpler win condition in a very static environment. The agent had to flounder around a lot until it bumped into the correct location for a few consecutive iterations. The environment is very static so once a solution is found it just has to memorize it. A2C has the exploration advantage and can find the solution sooner but it keeps exploring and does not converge to the single conclusion as quickly.

Missile Command shows a similar performance for the three algorithms. Although Prioritized Dueling DQN finds a higher value in earlier stages, The three algorithms get trapped in a local optimum. In the game missile command, four fire-balls target three bases. To get all 8 points the player has to defend all three. One of the bases gets attacked by two fire-balls which make it hard to defend. To have time to save the third base requires very accurate play, the agents did not seem to be able to maintain a perfect score because a few missteps led to 5 points. The reward plain is very non-linear for this game.

Superman takes this difficulty to the next level. The game is very dynamic with many NPCs modifying the environment in a stochastic manner. This means that any actions that the agent takes will have a big impact on the environment in the future. On top of this, the way to get the most points is to capture the antagonists and take them to jail. No points are awarded for capture, only for delivery to jail. This introduces a delayed reward which is a barrier to discovery. Knowing this, the results from the training on this game make sense. The agents were occasionally able to stumble on a good pattern but they could not reproduce the success in the stochastic environment.  

DQN and Prioritized Dueling DQN struggled to play Boulder Dash. In Boulder Dash, when the player collects a diamond for points, a rock falls toward them. This means there is negative feedback if an agent collects a diamond and doesn't move. Not collecting any diamonds and surviving appears to be an obvious local optimum that the agents have a hard time escaping. On the other hand, A2C was able to discover how to collect diamonds and survive, with a clear trend of continuing to improving. 

Seaquest is a good example of a game that is not too hard but has a lot of random elements. The agent can get a high score if it can survive the randomly positioned fish, catch the randomly moving diver, and take it to the surface. This requires the agent to learn to chase the diver which none of the agents appear to be doing. The high noise in the results is most likely from the agents failing to learn the general rules behind the stochasticity. Additionally, the player needs to go to the surface every $25$ game ticks or it loses the game, which may be something hard to learn for the agents.

Finally, Zelda is a fairly good game for reinforcement learning. Though, the game is not too similar to its namesake. The player must find a key and use it to unlock the exit while fighting enemies. Each event provides feedback which allows the agents to learn the game well.

%-------------------------------------------------------------
%-------------------------------------------------------------
%-------------------------------------------------------------

\subsection{Comparison with ALE}

Reinforcement learning research has been making a lot of progress on game playing in the last few years and the benchmark environments need to keep up. ALE is a popular 2D environment. It consists of a reasonably large set of real games and all the games have been designed for humans. Yet, the game set is static and cannot provide new challenges as researchers experiment with the strengths and weaknesses of different algorithms.

GVGAI currently has over twice the number of games as ALE and with active research more are added every year. The VGDL language also makes it possible for researchers to design new games. Truly stochastic games can be designed and multiple levels can be included to test how well an algorithm can generalize. The VGDL engine also provides a forward model that can be incorporated in the future to allow hybrid algorithms to learn and plan.  

While these games allow targeted testing of AIs, they tend to not be designed with humans in mind and can be hard to play. Readers are also not as familiar with the games as they are in Atari and therefore might lack some of the intuition. Another drawback is speed. The engine is written in Java and communicating through a local port to Python. While still very fast, training will run a few times slower than Atari. Currently, there is ongoing development to optimize the communication between the two languages.     

While both environments share some games, the performance on these games cannot be compared directly. GVGAI has games that are inspired by Atari but they are not perfect replicas and the author of the VGDL file can decide how close to match the original and how to handle score.
Yet, looking at similar games in both environments seems to show that GVGAI can have many of the characteristics of Atari: such as fairly good performance on Aliens and poor performance on Seaquest.

The ALE has done a lot for providing a standard benchmark for new algorithms to be tested against. GVGAI is more fluid and changing but it allows researchers to constantly challenge the perceived success of new RL agents. The challenges for computers can advance with them all the way to general video game playing. On top of that, we provide the results here to propose that doing well on GVGAI is at least comparable doing well on ALE and we show that there are games on GVGAI that still are not beaten.

\subsection{Comparison with planning algorithms}

In order to compare the performance of our learning algorithms with the state-of-art, we have used the results obtained in~\cite{bontrager2016matching}. This paper explores clustering GVGAI games to better understand the capabilities of each algorithm and subsequently use several agents to test the performance of each representative game. The tested agents may be classified in Genetic Algorithms (GA), Monte Carlo Tree Search (MCTS), Iterative With and Random Sample (RS). To compare results, we took the agent with the high score for each category in a target environment.

In Table \ref{tab:planning} we compare the performance of the reinforcement-learned neural network agents with high-performing planning agents. This is very much a case of comparing apples and oranges: the learning-based agents have been trained for hours for the individual game it is being tested on whereas the planning-based agents have had no training time whatsoever and are supposed to be ready to play any game at any point, and the planning-based agents have access to a forward model which the learning agent does not. In other words, each type of agent has a major advantage over the other, and it is a priori very hard to say which advantage will prove to be the most important. This is why this comparison is so interesting.

Beginning with Aliens, we see that all agents learn to play this game well. This is not overly surprising, as all Non-player Characters (NPC) and projectiles in this game behave deterministically (enemy projectiles are fired stochastically, but always takes some time to reach the player) and the game can be played well with very little planning; the main tasks are avoiding incoming projectiles and firing at the right time to hit the enemy. The former task can be solved with a reactive policy, and the latter with a minimum of planning and probably also reactively.

Wait for Breakfast was solved perfectly by all agents except the standard MCTS agent, which solved it occasionally. This game is easily solved if you plan far enough ahead, but it is also very easy to find a fixed strategy for winning. It punishes ``jittery'' agents that explore without planning.

Frogs is only won by the planning agents (GA and IW always win it, MCTS sometimes wins it) whereas it is never won by the learning algorithm. The simple explanation for this is that there are no intermediate rewards in Frogs; the only reward is for reaching the goal. There is, therefore, no gradient to ascend for the reinforcement learning algorithms. For the planning algorithms, on the other hand, it is just a matter of planning far enough ahead. (Some planning algorithms do better than others, for example, Iterative Width looks for intermediate states where facts about the world have changed.) The reason why learning algorithms perform well on Freeway, the Atari 2600 clone of Frogger, is that it has plenty of intermediate rewards - the player gets a score for advancing each lane.

Two of the planning agents and all three learning agents perform well on Missile Command; there seems to be no meaningful performance difference between the best planning algorithms (IW) and the learning agents. It seems possible to play this game by simply moving close to the nearest approaching missiles and attacking it. What is not clear is why MCTS is performing so badly.

Seaquest is a relatively complex game requiring both shooting enemies, rescuing divers and managing oxygen supply. All agents play this game reasonably well, but somewhat surprisingly, the learning agents perform best overall and A2C is the clear winner. The presence of intermediate rewards should work in the learning agents' favor; apparently, the learning agents easily learn the non-trivial sequence of tasks as well.

Boulder Dash is perhaps the most complex game in the set. The game requires both quick reactions for the twitch-based gameplay of avoiding falling boulders and long-term planning of in which order to dig dirt and collect diamonds so as not to get trapped among boulders. Here we have the interesting situations the one planning algorithm (MCTS) and one learning algorithm (A2C) plays the game reasonably well, whereas the other algorithms (both planning and learning) perform much worse. For the planning algorithms, the likely explanation is that GA has too short planning horizon and IW does not handle the stochastic nature of the enemies.

For Zelda, which combines fighting random-moving enemies and finding paths to keys and doors (medium-term planning), all agents performed comparably. The tree search algorithms outperformed the GA, and also seem to outperform the learning agents, but not by a great margin.
 
%-------------------------------------------------------------
\begin{table*}[!t]
\centering
\setlength\tabcolsep{1.5pt}
\begin{tabular}{|c|c|ccc|ccc|}
\hline
\multirow{1}{*}{\bff{Games}} &\multirow{1}{*}{\bff{Random Agent}} &\multicolumn{3}{c|}{\bff{Planning Agents}} &\multicolumn{3}{c|}{\bff{Learning Agents}}\\
& & \bff{Genetic Algorithm}& \bff{Monte Carlo Tree Search} & \bff{Iterative Width} & \bff{DQN} & \bff{Prioritized Dueling DQN} & \bff{A2C}   \\
 \hline
 Aliens & 52 & \textbf{80.4} & 72.6 & 80.2 & 75 & 74 & 77\\
 Wait For Breakfast & 0 & \textbf{1} & 0.4 & \textbf{1} & \textbf{1} & \textbf{1} & \textbf{1}\\  
 Frogs & -2 & \textbf{1} & -0.4 & \textbf{1} & 0 & 0 & 0\\  
 Missile Command & -2.2 & 2.6 & -3 & 6.8 & 5 & \textbf{8} & 5\\   
 Seaquest & 17.2 & 435 & 638.2 & 224.6 & 600 & 800 & \textbf{1200}\\   
 Boulder Dash & 1.4 & 3.4 & \textbf{16.4} & 8.8 & 2.5 & 5 & 15.5\\ 
 Zelda &-5.2 &3.4 &6.8 & \textbf{7.6}&4.2 &4.2 &6\\
 Superman &4 &157 &\textbf{6699} &130.2 &500 &0 &800\\ 
\hline
\end{tabular}
\caption{Learning score comparison of learning algorithms (DQN, Prioritized Dueling DQN and A2C) with random and planning algorithms (Genetic Algorithms, MCTS and Iterative Width). The results of planning and random are taken from \cite{bontrager2016matching} and correspond to the best performing instance of each algorithm.\\
}
\label{tab:planning}
\end{table*}
%-------------------------------------------------------------

\section{Conclusion}

In this paper, we have created a new reinforcement learning challenge out of the General Video Game AI Framework by connecting it to OpenAI Gym environment. We have used this setup to produce the first results of state-of-art deep RL algorithms on GVGAI games. Specifically, we tested DQN, Prioritized Dueling DQN and Advance Actor-Critic (A2C) on eighth representative GVGAI games. 

Our results show that the performance of learning algorithm differs drastically between games. In several games, all the tested RL algorithms can learn good stable policies, possibly due to features such as memory replay and parallel actor-learners for DQN and A2C respectively. A2C reaches a higher score than DQN and PDDQN for 6 of the 8 environments tested without memory replay. Also, when trained on the GVGAI domain using 12 CPU cores, A2C trains five times faster than DQN trained on a Tesla Nvidia GPU.

But there are also many cases where some or all of the learning algorithms fail. In particular, DQNs and A2C perform badly on games with a binary score (win or lose, no intermediate rewards) such as Frogs. Also, we observed a high dependency of the initial conditions which suggests that running multiple times is necessary for accurately benchmarking DQN algorithms. Finally, some complex games (e.g. Seaquest) show problems of stabilization when we are training with default parameters of OpenAI baselines. This reflects that a modification of replay memory or the schedule of the learning rate parameters are necessary to improve convergence in several environments.  

We also compared learning agents (which have time for learning but not a forward model) with planning agents (which get no learning time, but do get a forward model). The results indicate that in general, the planning agents have a slight advantage, though there are large variations between games. The planning agents seem better equipped to deal with making decisions with a long time dependency and no intermediate rewards, but the learning agents performed better on e.g. Seaquest (a complex game) and Missile Command (a simple game).

As researchers experiment with more the existing games, design specific games for experiments, and participate in the competition, we expect to gain new insights into the nature of various learning algorithms. There is an opportunity for new games to be created by humans and AIs in an arms race against improvements from game-playing agents. We believe this platform can be instrumental to scientifically evaluating how different algorithms can learn and evolve to understand many changing environments.

%The next steps are to improve the quality of the results showed for the learning agent: i) increase the number of simulations for training phase ii) increase the number of games of GVGAI tested iii) increase the number of learning algorithms used to validate OpenAI baseline library iv) run multiple times each learning algorithm for each target environment V) testing algorithm in a play mode to validate results. 

\section*{Acknowledgement}
This work was supported by the Ministry of Science and Technology of China (2017YFC0804003).

(*) The first two authors contributed equally to this work.

%\bibliographystyle{alpha}
%\bibliography{sample}
\bibliographystyle{IEEEtran}
\bibliography{./bibs/jliu,./bibs/gameseminar,./bibs/qmul,./bibs/planetwars,./bibs/jialinbib,./bibs/bibliography}

\end{document}